\documentclass{article}
\usepackage{acro_preprint,times}
\usepackage[T1]{fontenc}

\usepackage{amsfonts, amsmath, amssymb, bbm}
\usepackage{xcolor}
\usepackage{colortbl}
\usepackage{booktabs}
\usepackage{wrapfig}
\usepackage{afterpage}
\usepackage{needspace}
\usepackage{placeins}
\usepackage{graphicx}
\usepackage{subcaption}
\usepackage{tikz}
\usepackage{algorithm}
\usepackage{algpseudocode}
\usepackage{hyperref}
\usepackage{url}
\definecolor{linkblue}{HTML}{2878B5}
\hypersetup{
  colorlinks=true,
  citecolor=linkblue,
  linkcolor=linkblue,
  urlcolor=linkblue,
  pdftitle={ACRO: Actor--Critic Rollout Orchestration},
  pdfsubject={Preprint},
  pdfauthor={Junhoo Lee, Seungyeon Kim, Baekseung Kim, Minkyu Kim, Suhyun Jeon, Jimyeong Kim, Nojun Kwak}
}
\usepackage[capitalise]{cleveref}

\makeatletter
\newcommand{\finishwrap}[1][1]{%
  \par
  \ifnum\c@WF@wrappedlines>#1\relax
    \@tempdima=\baselineskip
    \multiply\@tempdima by \c@WF@wrappedlines
    \advance\@tempdima by -#1\baselineskip
    \vskip\@tempdima
  \fi
  \WFclear
}
\makeatother

\newcommand{\acrorevision}[1]{#1}

\title{\acrorevision{ACRO: Actor--Critic Rollout Orchestration}}

\author{
Junhoo Lee$^{1}$ \quad Seungyeon Kim$^{2}$ \quad Baekseung Kim$^{2}$ \quad Minkyu Kim$^{2}$ \\
\bfseries Suhyun Jeon$^{2}$ \quad Jimyeong Kim$^{2}$ \quad Nojun Kwak$^{2}$ \\[4pt]
\normalfont $^{1}$Korea Advanced Institute of Science and Technology (KAIST) \\
\normalfont $^{2}$Seoul National University
}

\begin{document}

\maketitle

\begin{abstract}
Executing a vision-language-action (VLA) policy involves deciding when to continue an attempt and when to intervene to improve its chances of success. We introduce ACRO, an Actor--Critic rollout orchestration layer that makes these decisions for a fixed VLA. ACRO seeks interventions that offer a more promising continuation, are physically feasible, and leave sufficient time to complete the task. A Critic learned from policy rollouts estimates success prospects to decide when to intervene and select a re-entry target along the executed trajectory. Trajectory Retraction shortcuts the visited path to return the robot to that target, where the same VLA resumes execution from a fresh observation. We evaluate ACRO with GR00T N1.7 and $\pi_{0.5}$ across SimplerEnv, RoboCasa, and real-robot manipulation. Under the same physical execution budget, including all return motion, ACRO improves success by 10.4, 7.3, and 21.7 percentage points, respectively. It also completes additional tasks as the execution budget grows. These results show that controlling when and where a policy resumes can turn its existing capabilities into more successful task executions. Project page: \url{https://junhoo.me/acro}.
\end{abstract}

\section{INTRODUCTION}

\afterpage{\begin{figure}[!t]
  \centering
  \includegraphics[width=\linewidth]{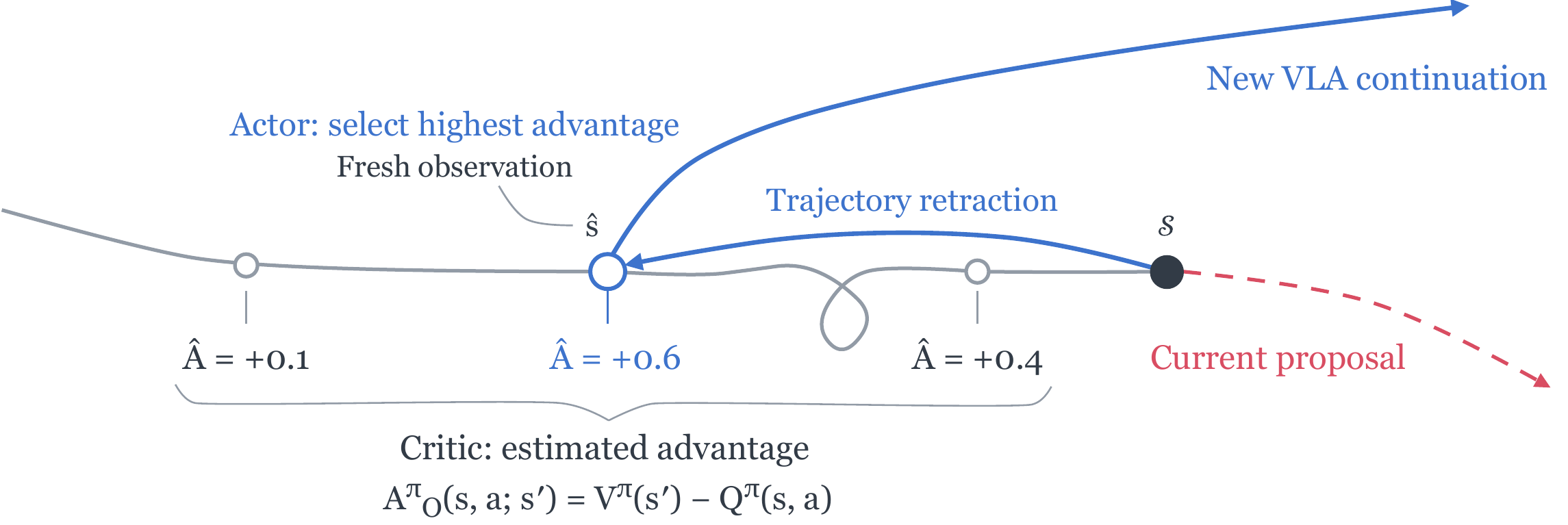}
  \caption{\textbf{Actor--Critic rollout orchestration.} The orchestration advantage $A^\pi_{\mathcal O}(s,a;s')=V^\pi(s')-Q^\pi(s,a)$ compares task-success probabilities after re-entry and after executing the current proposal, within the available time. Once intervention is activated, the Actor selects the recorded candidate with the highest estimated advantage; the current $Q$ is fixed during this comparison. Trajectory Retraction shortens the return path, and the VLA resumes from a fresh observation.}
  \label{fig:acro-overview}
\end{figure}
}

Teaching a robot to act leaves open the question of when to trust its policy to continue and when to intervene. Vision-language-action (VLA) models map visual observations and language instructions to robot actions~\citep{zitkovich2023rt2,kim2025openvla,black2025pi0}. These models can succeed on a task in one execution and fail on the same task in another, suggesting that the outcome of a single execution may fall short of the policy's capabilities. Orchestration that translates this potential into successful task completion could improve the performance of an existing VLA without retraining the policy.

Effective orchestration requires two decisions: when to intervene, and how far that intervention should go. RoboMonkey and CoVer evaluate and select action proposals from the current observation~\citep{kwok2025robomonkey,kwok2026cover}, while AutoHorizon adjusts how much of a predicted action chunk to execute~\citep{wang2026autohorizon}. These decisions improve the policy's next execution, but evaluating an unfavorable proposal does not itself specify a physical intervention that would enable a better continuation. External orchestrators such as Pigey organize subgoals and skill calls, verify their outcomes, and apply retry and fallback rules~\citep{galanti2026pigey}. In Pigey's VLA interface, however, the orchestrator reassesses execution after a bounded rollout returns. Such verification and routing rules do not directly determine when to interrupt an ongoing VLA execution or how much to change its state before returning control. Intervening only as much as is useful requires evaluating the policy's prospects both if the current action proceeds and if execution resumes after an intervention.

We introduce ACRO, an Actor--Critic rollout orchestration layer that makes these decisions according to the deployed VLA's probability of task completion. ACRO learns action- and state-value estimates from execution outcomes. The Critic evaluates the VLA's proposed action to determine when to interrupt execution. To turn this assessment into a tractable intervention, we introduce \emph{Trajectory Retraction}, which reconstructs the observed execution path into a constrained region for intervention search. Within this region, ACRO uses state values to select a promising target reachable under the intervention budget. The Actor moves the robot to the selected configuration and returns control to the VLA, which observes the resulting state and generates a new continuation. Repeating this process within the remaining budget allows a minimal intervention layer to support alternative successful executions while leaving task execution to the existing policy.

\acrorevision{We evaluate ACRO in simulation and on real-robot manipulation tasks. ACRO improves task success under the same execution budget, even when the physical steps used for trajectory retraction are included. These gains extend to the real robot, where success increases from $43.3\%$ to $65.0\%$. As the execution budget grows, ACRO achieves further gains in task completion, demonstrating test-time scaling through rollout orchestration.}

\acrorevision{Our contributions are as follows:}
\begin{itemize}
\setlength{\itemsep}{2pt}
\item \acrorevision{We formulate rollout orchestration through the difference in task-success probability between continuing a proposed action and resuming the policy after a physical intervention.}
\item \acrorevision{We introduce ACRO, which uses a shared Q/V Critic to decide when to intervene and Trajectory Retraction to select and reach a promising continuation point.}
\item \acrorevision{We show that ACRO improves success with fixed VLAs on SimplerEnv, RoboCasa, and real robots under matched physical execution budgets that include retraction.}
\end{itemize}

\section{RELATED WORK}

\paragraph{Robot architectures and orchestration.}
Robot architectures such as CLARAty, CRAM, and hierarchical planning in the now connect planning to execution through runtime feedback and adjustment~\citep{volpe2001claraty,beetz2010cram,kaelbling2011now}. More recent systems organize learned robot capabilities: SayCan grounds skill selection in affordance values~\citep{ichter2023saycan}, Inner Monologue incorporates execution feedback into language planning~\citep{huang2023innermonologue}, and Pigey coordinates subgoals, skill calls, and retries~\citep{galanti2026pigey}. These mechanisms support task-level coordination, but do not directly specify when to interrupt an ongoing policy action or how much physical intervention would improve its continuation.

\paragraph{Test-time control of VLA execution.}
Test-time methods improve deployed VLAs by controlling action selection and execution. V-GPS and RoboMonkey use learned evaluators to select action proposals~\citep{nakamoto2025steering,kwok2025robomonkey}, while CoVer jointly selects instructions and action chunks through alignment scoring~\citep{kwok2026cover}. AutoHorizon adjusts how much of a predicted chunk to execute before replanning~\citep{wang2026autohorizon}. These methods refine execution from the current observation, leaving open how to select a physical intervention when continuing from that state offers poor prospects for success.

\paragraph{Failure detection and recovery.}
Runtime failure detection provides signals for intervention. SAFE predicts failure from VLA features~\citep{gu2025safe}, while FIPER combines observation novelty with action uncertainty~\citep{romer2025fiper}. Such signals alone do not determine where the policy should resume. CycleVLA combines a progress-aware VLA with an external VLM that predicts failures and selects subtasks for backtracking, followed by minimum Bayes risk decoding for retries~\citep{ma2026cyclevla}. Recovery decisions therefore depend on an external VLM's assessment of subtask preconditions, whose connection to the deployed policy's actual ability to recover remains unclear. Moreover, organizing intervention around subtask boundaries makes its timing depend on the chosen task decomposition; these boundaries need not coincide with the moments when intervention would most benefit execution.

\section{Method}

\subsection{Problem \acrorevision{Formulation}}

\paragraph{Goal} Our goal is to improve a VLA's task success through orchestration. To be useful in real robotic deployments, orchestration must account for both the time required to intervene and the physical feasibility of the intervention. Meaning, interventions must be physically executable and timely: assessing the ongoing execution, selecting a re-entry target, and moving the robot must leave sufficient time for task completion. We therefore seek interventions that enable more promising policy continuations within the available time.

\paragraph{Objective}
Our goal can be viewed as finding an intervention whose resulting
continuation offers an advantage over executing the currently
proposed action. Let $s$ denote the available execution context,
including the observation history, task instruction, and elapsed
time, and let $a\sim\pi(\cdot\mid s)$ denote the action proposed by
the VLA. We define $Q^\pi(s,a)$ as the probability of task completion
within the time limit after executing $a$ and then following $\pi$,
and $V^\pi(s)$ as the corresponding probability when $\pi$ selects
its continuation from $s$, following the standard state- and action-value definitions~\citep{sutton2018reinforcement}. For an intervention that reaches $s'$, including the elapsed time
at policy resumption, we define its advantage over executing $a$ as
\begin{equation}
A^\pi_{\mathcal O}(s,a;s')
=
V^\pi(s')-Q^\pi(s,a).
\label{eq:orchastration-advantage}
\end{equation}
Here, $\mathcal O$ denotes an orchestration decision that changes where the VLA resumes execution.

\paragraph{Challenges}
However, realizing this advantage at test time is not straightforward. We must estimate whether continuing with the proposed action is likely to succeed. If the prospects are low, we need to find a state $s'$ that offers a better continuation. Searching more broadly may uncover better targets, but takes time and can lead to states where value estimates are less reliable. We must then move the robot toward the selected target and leave enough time to complete the task. The scene may also change during the move, affecting the continuation we expected.

\subsection{ACRO}
\label{sec:acro}
\paragraph{Actor--Critic Orchestration.}
We implement the objective in Eq.~\ref{eq:orchastration-advantage}
through an Actor--Critic structure. The Critic evaluates the
success prospects of policy continuations, and the Actor uses
these evaluations to select and execute a repositioning motion.
Directly evaluating the orchestration advantage requires searching
for alternative targets and assessing their continuation values.
Doing so at every policy step would spend time even when the
current execution is likely to succeed. We therefore adopt a
sparse Actor--Critic structure: a Q-based assessment identifies
when to activate the Actor, which then searches for a promising
target using V. Once activated, the current action value is fixed,
so maximizing V over candidate targets also maximizes their
estimated orchestration advantage.

\paragraph{Evaluating Continuations.}
Repositioning and retries consume time that could otherwise be
used to complete the task. A useful continuation must therefore
offer a high probability of success within the available time.
We learn Q and V from rollouts of the deployed VLA by modeling
the time to task completion. Q conditions on the observation
history and proposed action, while V uses the observation history
alone. Both predict completion hazard rates, yielding success
probabilities over a forecast horizon $h$:
\begin{equation}
\widehat V(s)=1-e^{-\int_0^h\lambda_V(u\mid s)\,du},
\qquad
\widehat Q(s,a)=1-e^{-\int_0^h\lambda_Q(u\mid s,a)\,du}.
\label{eq:time-to-success-values}
\end{equation}
Here, $\lambda_V$ and $\lambda_Q$ are the predicted completion
rates conditional on the task not yet being completed. We
parameterize them as piecewise-constant rates over time intervals~\citep{kvamme2021survival}.
\acrorevision{The forecast horizon $h$ is the interval over which the Critic predicts success; the episode budget counts all physical steps taken by the policy and by retraction.}

These predictions must account for both observed successes and
executions that end before success is observed. A successful
rollout provides a completion time; an incomplete rollout indicates
only that completion did not occur during the observed interval.
We incorporate both through the censored time-to-success loss
\begin{equation}
\ell_{\mathrm{surv}}(\lambda;d,\delta)
=\int_0^d\lambda(u)\,du-\delta\log\lambda(d),
\label{eq:survival-loss}
\end{equation}
where $d$ is the duration from a policy decision to completion or
observation cutoff, and $\delta$ indicates whether completion
was observed. We apply this loss to both Q and V.

The value of a proposed action should also agree with the
continuation it produces. For an action interval of duration
$\Delta t$ without termination, the finite-horizon Bellman relation gives~\citep{sutton2018reinforcement}
\begin{equation}
Q_h^\pi(s,a)
=\mathbb E\!\left[V_{h-\Delta t}^\pi(s')\mid s,a\right].
\label{eq:qv-consistency}
\end{equation}
We encourage this consistency by additionally training Q toward
a stop-gradient V prediction at the next observed state, reducing
the forecast horizon by the elapsed action duration. Observed
success in the final action interval supplies a target of one.
Censored endpoints without a subsequent observation are excluded
from this additional bootstrap term.

Before target search, the alternative continuation value is
unavailable. We therefore use Q to predict whether the current
execution will fail within the task horizon. With the Critic
fixed, we train a logistic classifier on recent Q estimates
using rollout-level failure labels. Its prediction determines
when to activate the Actor. This provides a proxy for identifying
opportunities for positive advantage while reserving target
search and physical repositioning for selected moments.

\paragraph{Value-Guided Repositioning.}
The Actor must find a promising target and reach it with enough
time left for task completion. Searching more broadly may reveal
better targets, but increases search time and exposes selection
to value errors at unfamiliar states. We therefore use configurations
recorded along the active execution trajectory as candidates.
These records provide observations and value estimates for target
selection, together with a path for guiding physical movement.

Among the eligible recorded contexts $\mathcal R$, the Actor
selects the target with the highest estimated V. The recorded
value serves as a proxy for the prospects of resuming from its
associated robot configuration. To reduce movement time, Trajectory
Retraction shortcuts the recorded path while keeping each segment
near the visited positions. The target-selection rule and shortcut
condition are
\begin{equation}
\hat s\in\operatorname*{arg\,max}_{s'\in\mathcal R}\widehat V(s'),
\qquad
\max_{p\in[p_i,p_j]}\operatorname{dist}(p,\mathcal P)\le\rho.
\label{eq:value-guided-repositioning}
\end{equation}
Here, $\mathcal P$ is the set of recorded end-effector positions
along the selected trajectory segment, and $\rho$ is the allowed
deviation. After target selection, we greedily choose the farthest
waypoint toward the target whose connecting segment satisfies
the condition, approximated by sampling points along the segment.
The Actor follows the resulting path, and the VLA receives a
fresh observation of the reached state to generate a new
continuation. Subsequent execution updates the active trajectory
for later orchestration decisions. Algorithm~\ref{alg:acro} summarizes the online orchestration loop.
\begin{algorithm}[t]
\caption{Actor--Critic Rollout Orchestration (ACRO)}
\label{alg:acro}
\small
\begingroup
\algrenewcommand{\algorithmiccomment}[1]{\hfill\textcolor{linkblue}{$\triangleright$ #1}}
\begin{algorithmic}
\State Initialize trajectory history $\mathcal B$, Q history $\mathcal Q$, and elapsed steps $n\gets0$
\While{$n<T$ and the episode has not terminated}
    \State Observe $s$ from the reached scene; sample $a\sim\pi(\cdot\mid s)$
    \State $q\gets\widehat Q(s,a)$, $v\gets\widehat V(s)$ \Comment{Evaluate continuations}
    \State Store $(s,v)$ and the robot pose in $\mathcal B$; append $q$ to $\mathcal Q$
    \If{$g(\mathcal Q)>\tau$ and re-entry is available}
        \State $\hat s\gets\operatorname*{arg\,max}_{s'\in\mathcal R}\widehat V(s')$
            \Comment{Select an eligible recorded target}
        \State $\mathcal P\gets$ recorded path from the current pose toward $\hat s$
        \State Shortcut $\mathcal P$ within distance $\rho$ of visited positions \Comment{Trajectory Retraction}
        \State Follow the shortened path; let $\Delta n$ be the physical steps used
    \Else
        \State Execute $a$ until the next query; let $\Delta n$ be the steps used
    \EndIf
    \State Update the active trajectory $\mathcal B$; $n\gets n+\Delta n$
\EndWhile
\end{algorithmic}
\endgroup
\footnotesize
The budget $T$ counts all physical steps; motion stops at termination or budget exhaustion.
Re-entry availability includes a nonempty
eligible set $\mathcal R$ and the intervention conditions in Appendix~\ref{app:trigger}--\ref{app:retraction}.
\end{algorithm}

\section{Experiments}
\label{sec:experiments}
We evaluate ACRO across pretrained policies, simulated environments, and a real robot. Section~\ref{sec:main_results} examines its use across these execution settings and the data needed to prepare the Critic. Section~\ref{sec:analysis} studies Critic-guided recovery and the efficiency of execution after returning. Section~\ref{sec:data-retraction} examines test-time scaling and physical return cost. Appendix~\ref{sec:ablations} examines Critic architectures and intervention triggers.

\Needspace{9\baselineskip}
\subsection{Experimental Setup}
\label{sec:exp_setup}

\begin{figure}[!t]
  \centering
  \begin{minipage}{0.88\linewidth}
    \captionsetup[subfigure]{font=small,skip=4pt,justification=centering}
    \begin{subfigure}{\linewidth}
      \centering
      \includegraphics[width=0.326\linewidth]{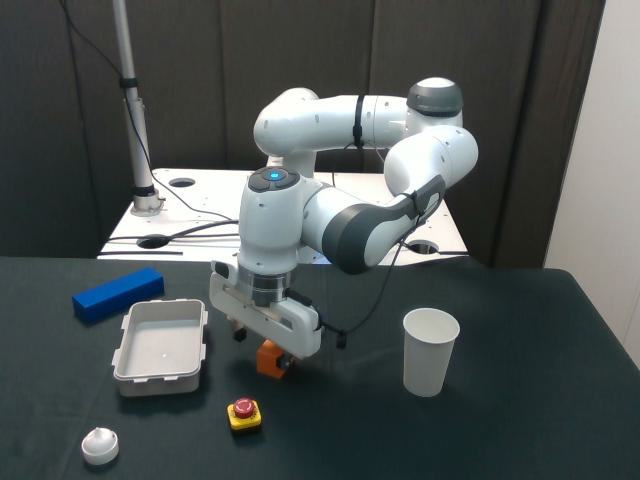}\hfill
      \includegraphics[width=0.326\linewidth]{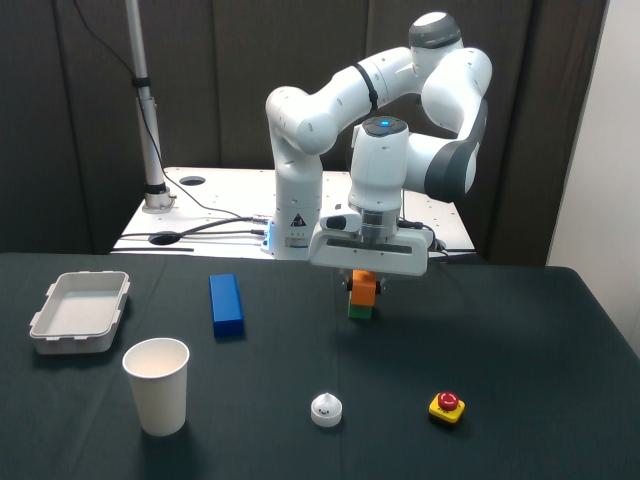}\hfill
      \includegraphics[width=0.326\linewidth]{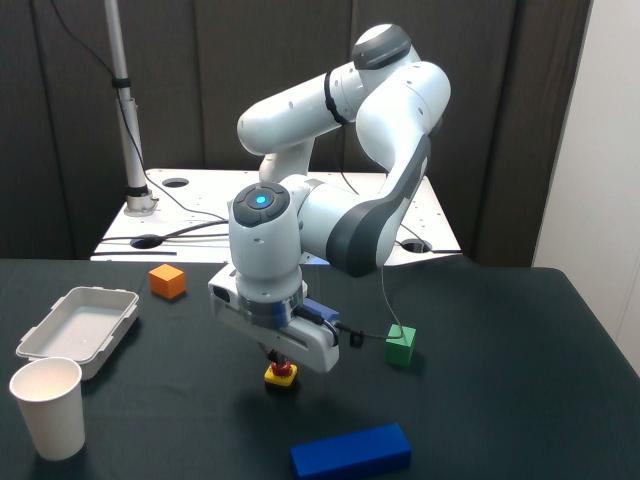}
      \caption{Real Robot}
      \label{fig:environments-real}
    \end{subfigure}

    \medskip
    \begin{subfigure}[t]{0.482\linewidth}
      \centering
      \includegraphics[width=0.489\linewidth,trim={240 0 240 0},clip]{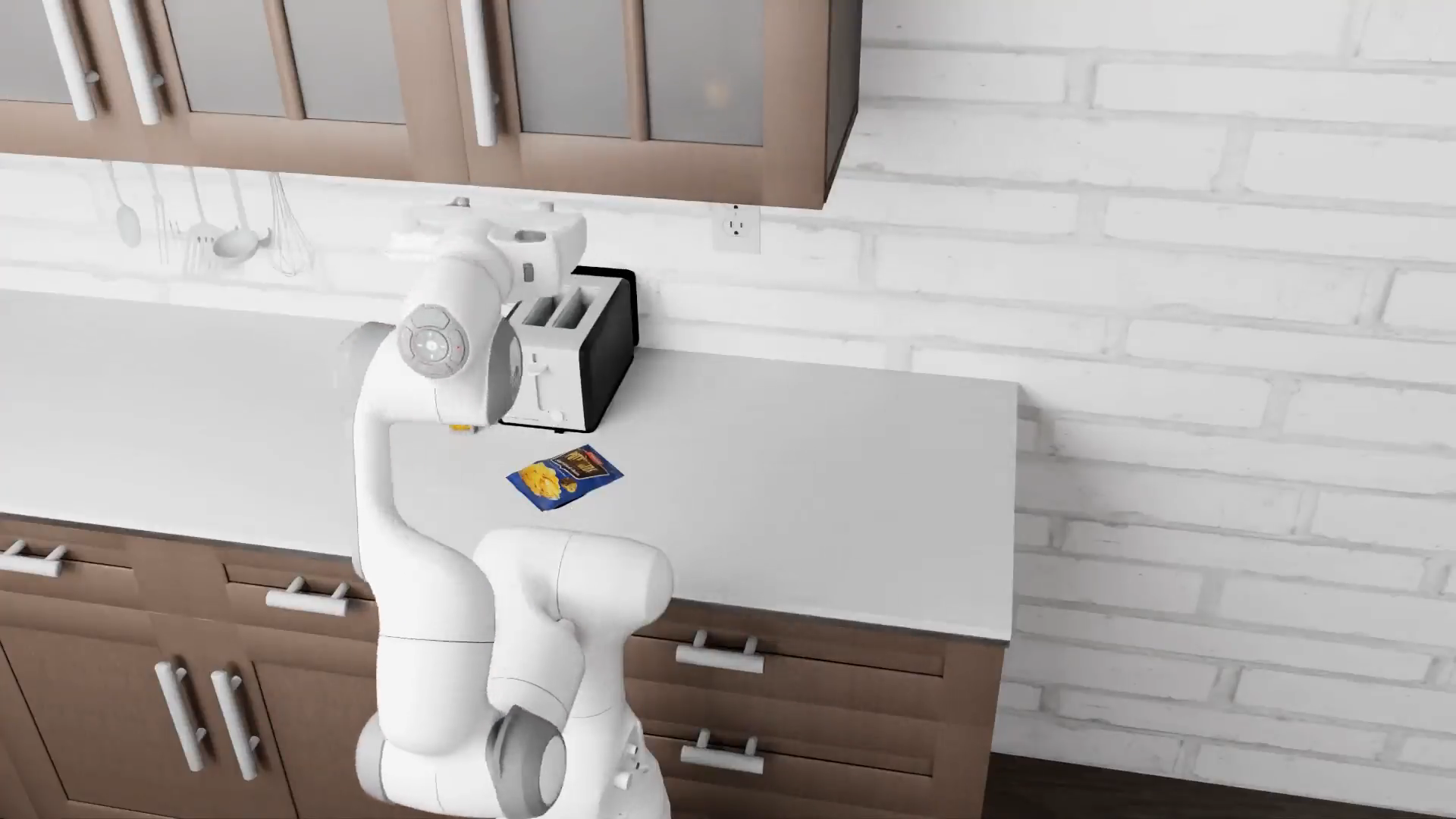}\hfill
      \includegraphics[width=0.489\linewidth,trim={350 0 130 0},clip]{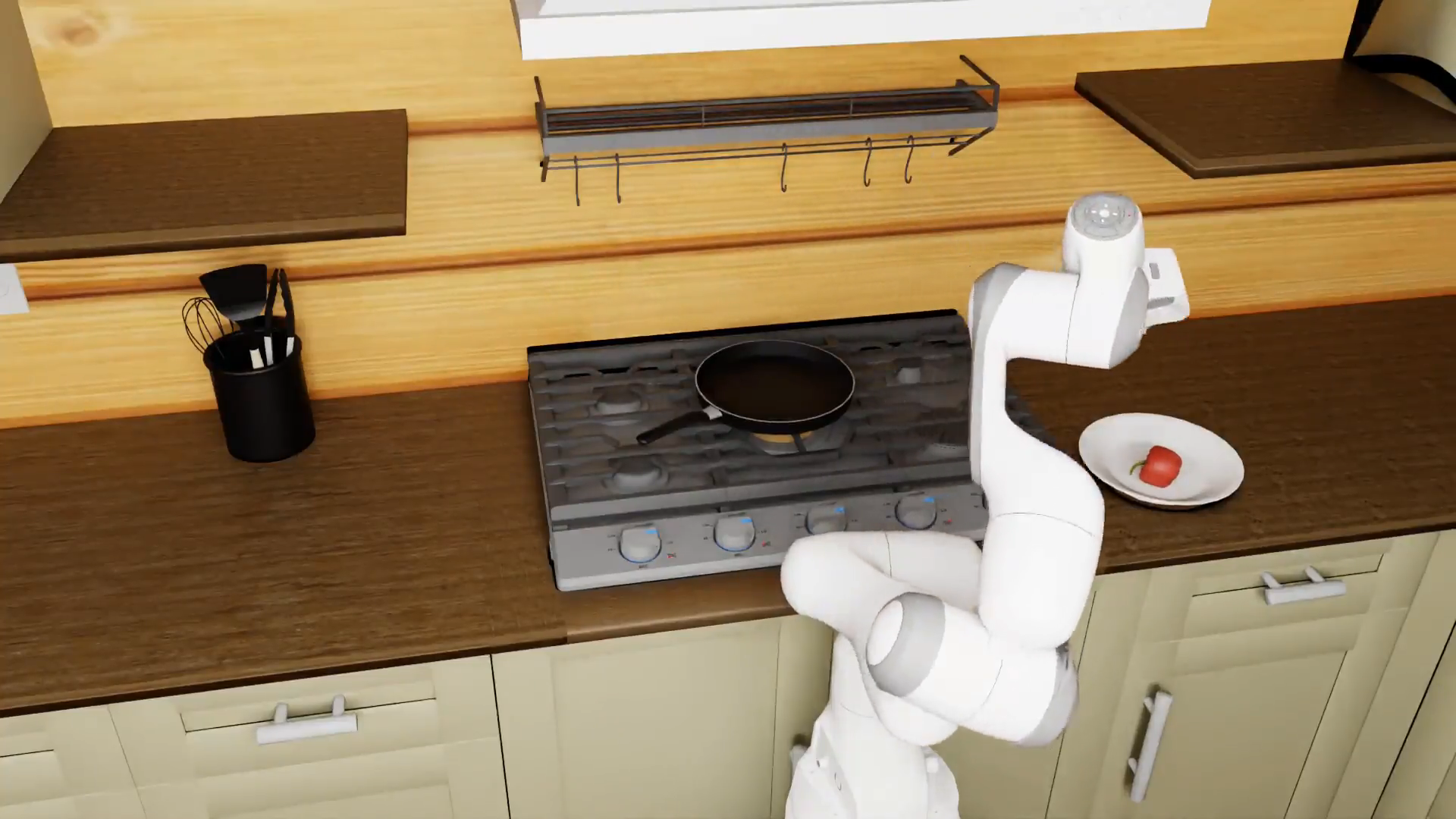}
      \caption{RoboCasa}
      \label{fig:environments-robocasa}
    \end{subfigure}\hfill
    \begin{subfigure}[t]{0.482\linewidth}
      \centering
      \includegraphics[width=0.489\linewidth,trim={20 62 20 0},clip]{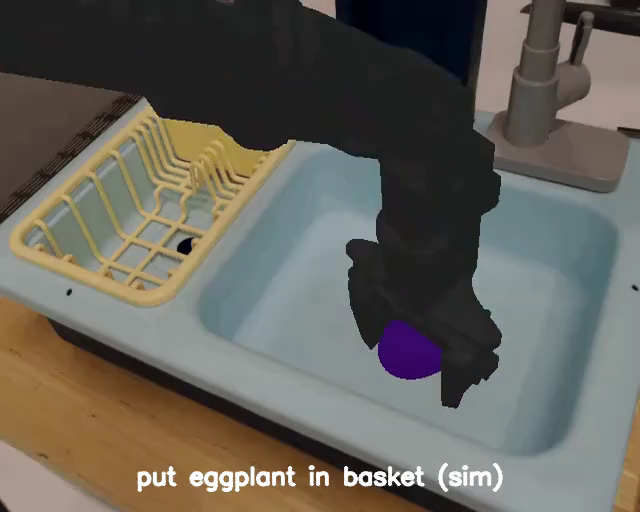}\hfill
      \includegraphics[width=0.489\linewidth,trim={20 62 20 0},clip]{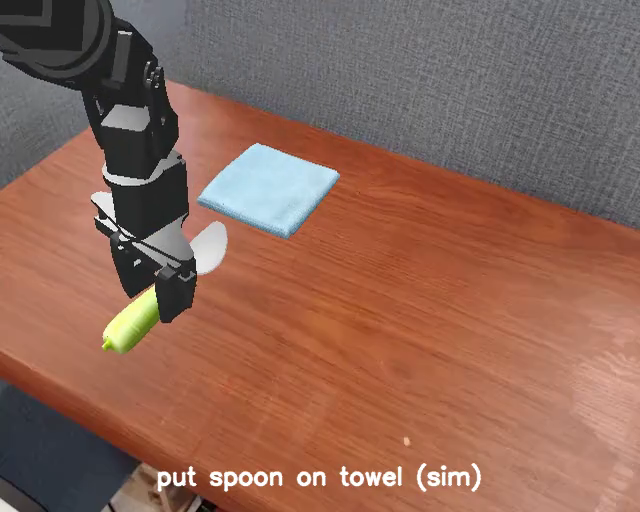}
      \caption{SimplerEnv Bridge}
      \label{fig:environments-bridge}
    \end{subfigure}
  \end{minipage}
  \caption{Evaluation environments. (a) Real robot: pick and place, cube stacking, and button pressing. (b) RoboCasa~\citep{nasiriany2024robocasa}: cabinet interaction and pick and place. (c) SimplerEnv Bridge~\citep{li2025simpler}: eggplant in basket and spoon on towel. Scenes are ordered left to right within each group; simulation images are from official project demonstrations.}
  \label{fig:environments}
\end{figure}

\paragraph{Models and Datasets.}
We use two pretrained VLAs with their policies fixed throughout execution: GR00T N1.7~\citep{nvidia2026groot17} on SimplerEnv Bridge~\citep{li2025simpler}, and $\pi_{0.5}$~\citep{black2025pi05} on RoboCasa~\citep{nasiriany2024robocasa}. SimplerEnv Bridge tests fine-grained tabletop manipulation, while RoboCasa covers a broader range of household object interactions. We also evaluate $\pi_{0.5}$ on a real robot to test recovery under physical sensing and actuation conditions. \Cref{fig:environments} introduces the three evaluation environments.

\paragraph{Baselines.}
\acrorevision{We compare three ways of using the execution budget: uninterrupted policy execution, value-ranked action selection, and recovery through physical repositioning.} Base VLA executes the underlying policy without intervention. RoboMonkey~\citep{kwok2025robomonkey} selects among candidate actions at the current state. \acrorevision{Our adaptation uses ACRO's Q estimate as its verifier to rank eight action chunks at each policy query. This compares using the same Critic for action selection with using it to guide ACRO's interventions, while requiring eight policy samples per query.} CycleVLA~\citep{ma2026cyclevla} uses a VLM to decompose the instruction into subtasks, check completion at subtask boundaries, and return the robot to the start of a failed subtask before retrying. \acrorevision{This provides a comparison with recovery guided by an external VLM. Our adaptation keeps the base policy fixed and omits the original policy fine-tuning.} On RoboCasa, Periodic rewind provides a fixed-schedule recovery baseline: it returns along the recorded path at fixed intervals to a fixed look-back distance. For real-robot experiments, we compare ACRO with Base VLA. \Cref{app:details} gives the implementation details of all methods.

\paragraph{Implementation Details.}
For a fair comparison, all methods use the same evaluation identities and receive the same physical step budget, which includes intervention motion. \acrorevision{Additional policy and VLM queries do not count toward this physical step budget.} We use each benchmark's standard episode limit: 300 steps on SimplerEnv, the official GR00T evaluation limit, and each task's registered limit on RoboCasa. To learn when and where to intervene, we train the Critic on uninterrupted rollouts of the base policy (1230 on SimplerEnv and 1200 on RoboCasa) using AdamW with a learning rate of $10^{-3}$, and set the intervention threshold by calibrating on held-out rollouts so that about 20\% of executions that would succeed raise an alarm. We measure task success over 240 trials on SimplerEnv, 600 on RoboCasa, and 60 on the real robot, pooling trials within each environment for overall performance. The scheduled-rewind and outcome analyses reuse all 600 RoboCasa evaluation trials at the same standard physical budget $H$. For the separate Critic and path ablations, we use $\pi_{0.5}$ on three RoboCasa tasks, one from each category: Turn On Electric Kettle (Control), Open Stand Mixer Head (Open), and Pick-and-Place Counter to Cabinet (Pick and Place).

\subsection{Across Policies and Execution Environments}
\label{sec:main_results}
\begin{table}[!t]
\centering
\small
\setlength{\tabcolsep}{6pt}
\caption{Task success rate (\%) in simulation. All methods run the same base policy under the benchmark's standard physical step limit, \acrorevision{including recovery motion}: 300 steps on SimplerEnv Bridge and each task's registered limit on RoboCasa. Overall pools all trials of a benchmark. Best result per row in bold. \acrorevision{$\dagger$Adapted baselines; see \cref{sec:exp_setup}.}}
\label{tab:main-results}
\begin{tabular}{@{}llrcccc@{}}
\toprule
 & & & \multicolumn{3}{c}{Baselines} & \\
\cmidrule(lr){4-6}
Policy & Task & $N$ & Base & \acrorevision{RoboMonkey$^\dagger$} & \acrorevision{CycleVLA$^\dagger$} & ACRO \\
\midrule
GR00T N1.7 & Carrot on Plate    & 60 & 50.0 & 31.7 & 48.3 & \textbf{68.3} \\
{\footnotesize SimplerEnv Bridge} & Eggplant in Basket & 60 & 36.7 & 50.0 & \textbf{66.7} & 51.7 \\
           & Spoon on Towel     & 60 & \textbf{93.3} & 90.0 & \textbf{93.3} & \textbf{93.3} \\
           & Stack Cube         & 60 & 18.3 & 18.3 & 13.3 & \textbf{26.7} \\
\cmidrule(l){2-7}
           & Overall            & 240 & 49.6 & 47.5 & 55.4 & \textbf{60.0} \\
\midrule
$\pi_{0.5}$ & Open/Close/Slide        & 250 & 64.4 & 66.8 & 65.6 & \textbf{76.0} \\
{\footnotesize RoboCasa} & Pick-and-Place & 250 & 64.0 & \textbf{68.0} & 56.0 & 65.2 \\
            & Control Operation       & 100 & 43.0 & 52.0 & 40.0 & \textbf{55.0} \\
\cmidrule(l){2-7}
            & Overall                 & 600 & 60.7 & 64.8 & 57.3 & \textbf{68.0} \\
\bottomrule
\end{tabular}
\end{table}

\paragraph{On Simulation Benchmarks.}
We evaluate ACRO across two pretrained VLAs and two simulation environments to test its use across different policies and manipulation settings. Table~\ref{tab:main-results} covers tabletop manipulation with GR00T~N1.7 on SimplerEnv Bridge and household interactions with $\pi_{0.5}$ on RoboCasa. ACRO raises overall success from 49.6\% to 60.0\% on SimplerEnv and from 60.7\% to 68.0\% on RoboCasa. It also exceeds RoboMonkey and CycleVLA overall in both settings. \acrorevision{RoboMonkey ranks eight policy samples per query with the same Critic, while CycleVLA uses external VLM inference for recovery.} With each base policy fixed and all return motion charged to the physical budget, these gains show that ACRO applies across distinct VLA backbones, environments, and task types.

\Needspace{14\baselineskip}
\begin{wraptable}{r}{0.47\linewidth}
\centering
\normalsize
\setlength{\tabcolsep}{5pt}
\caption{Real-robot success rate (\%).}
\label{tab:real-robot}
\vspace{2pt}
\begin{tabular}{@{}lrcc@{}}
\toprule
Task & $N$ & Base & ACRO \\
\midrule
Pick and Place & 20 & 70.0 & 80.0 \\
Stack Cube     & 20 & 20.0 & 55.0 \\
Press Button   & 20 & 40.0 & 60 \\
\midrule
Overall        & 60 & 43.3 & 65 \\
\bottomrule
\end{tabular}
\end{wraptable}

\paragraph{Real-Robot Evaluation.}
Physical retraction must bring the robot back to a useful configuration despite sensing noise and actuation error. We test this requirement on a real robot using $\pi_{0.5}$ for pick and place, cube stacking, and button pressing (Table~\ref{tab:real-robot}). ACRO improves success on all three tasks, with the largest gain on cube stacking (20.0\% to 55.0\%). \acrorevision{Across 60 trials, it completes 39 tasks compared with 26 for Base VLA, raising success from 43.3\% to 65.0\%.} The improvement under real-world execution conditions shows that physical return and renewed policy execution can be used together for successful recovery.
\finishwrap

\Needspace{20\baselineskip}
\begin{wrapfigure}{r}{0.48\linewidth}
\vspace{-4pt}
\captionsetup{justification=raggedright,singlelinecheck=false}
\centering
\includegraphics[width=\linewidth]{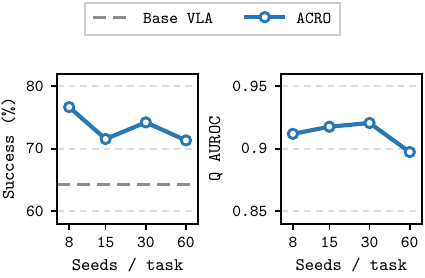}
\caption{\textbf{A small rollout set suffices to train the Critic.} Success and AUROC of minimum rollout Q over three tasks at $2H_{\mathrm{ref}}$.}
\label{fig:data-scale}
\end{wrapfigure}

\paragraph{Critic Training Cost.}
\label{sec:critic-data}
The Critic is trained from rollouts of the deployed policy. To measure the data needed to build it, we vary the training set from 8 to 60 seeds per task across 12 tasks, training one Critic with joint Q and V heads per size and fixing the validation and calibration sets at 180 and 300 rollouts. We evaluate on three RoboCasa tasks with 150 trials per task at $2H_{\mathrm{ref}}$ (\cref{app:protocol}). Figure~\ref{fig:data-scale} shows that ACRO exceeds Base VLA at every tested size. With 8 seeds per task, it reaches 76.7\% success using 96 training rollouts; AUROC ranges from 0.897 to 0.921 across sizes. These results show that a small training set is sufficient to build a Critic that improves task completion.
\finishwrap[3]

\Needspace{23\baselineskip}
\subsection{Critic-Guided Recovery}
\label{sec:analysis}
The preceding results show that ACRO improves completion across the evaluated policies and environments. We now examine the role of its recovery strategy, the successes gained through intervention, and the execution that follows a return.

\Needspace{16\baselineskip}
\begin{wraptable}{r}{0.49\linewidth}
\vspace{-4pt}
\captionsetup{justification=raggedright,singlelinecheck=false}
\centering
\small
\setlength{\tabcolsep}{4pt}
\caption{\textbf{ACRO improves on scheduled rewinds.} All 600 RoboCasa trials at $H$; returns and return steps are trial means.}
\label{tab:periodic-rewind}
\begin{tabular}{@{}lrrr@{}}
\toprule
Method & Success (\%) & Returns & Steps \\
\midrule
Base VLA & 60.7 & 0 & 0 \\
Periodic & 59.5 & 1.04 & 92.7 \\
\rowcolor{linkblue!12}
ACRO & 68.0 & 1.06 & 46.3 \\
\bottomrule
\end{tabular}
\end{wraptable}

\paragraph{Scheduled Rewinds.}
A fixed rewind schedule also gives the policy repeated attempts. We compare this strategy with ACRO to examine the benefit of using Critic estimates to guide recovery. Periodic rewind uses fixed intervals, fixed look-back distances, and reverse replay, while ACRO uses Q to activate recovery and V to select a target reached through Trajectory Retraction. On the same 600 RoboCasa trials at the standard physical budget $H$, Periodic completes 59.5\% of trials, compared with 60.7\% for Base VLA and 68.0\% for ACRO (Table~\ref{tab:periodic-rewind}). Periodic and ACRO initiate similar numbers of returns per trial (1.04 and 1.06), while ACRO uses about half as many return steps (46.3 versus 92.7). The Critic-guided strategy thus turns a similar number of recovery attempts into more completed tasks with less return motion.
\finishwrap

\Needspace{15\baselineskip}
\begin{wraptable}{r}{0.53\linewidth}
\vspace{-4pt}
\captionsetup{justification=raggedright,singlelinecheck=false}
\centering
\small
\setlength{\tabcolsep}{4.5pt}
\caption{\textbf{Gained successes exceed losses.} All 600 RoboCasa seeds at $H$, relative to Base. Policy-step means use the 46 shared successes after ACRO's first alarm.}
\label{tab:recovery-summary}
\begin{tabular}{@{}lrrrr@{}}
\toprule
 & \multicolumn{2}{c}{Outcome changes} & \multicolumn{2}{c}{Mean rollout steps} \\
\cmidrule(lr){2-3}\cmidrule(l){4-5}
Category & Gained & Lost & Base & ACRO \\
\midrule
Open/close & 33 & 4 & 362.6 & 296.1 \\
Pick/place & 17 & 14 & 387.0 & 345.5 \\
Control & 20 & 8 & 139.3 & 157.7 \\
\midrule
\textbf{Overall} & \textbf{70} & \textbf{26} & \textbf{327.9} & \textbf{287.3} \\
\bottomrule
\end{tabular}
\end{wraptable}

\paragraph{Gained and Lost Successes.}
An intervention can recover a failed execution, but it can also interrupt one that would succeed. To examine this balance, Table~\ref{tab:recovery-summary} decomposes the outcome differences over all 600 RoboCasa task--seed identities at $H$. ACRO succeeds on 70 seeds where Base fails and fails on 26 where Base succeeds, giving 44 additional completions. Periodic rewind gains 50 successes and loses 57. ACRO's net gain is positive in every category: +29 on open/close, +12 on control, and +3 on pick and place. Thus, the overall improvement reflects more gained than lost successes across the three categories. The comparison protocol is detailed in Appendix~\ref{app:recovery-timing}.
\finishwrap

\Needspace{8\baselineskip}
\begin{figure}[!t]
\centering
\includegraphics[width=\linewidth]{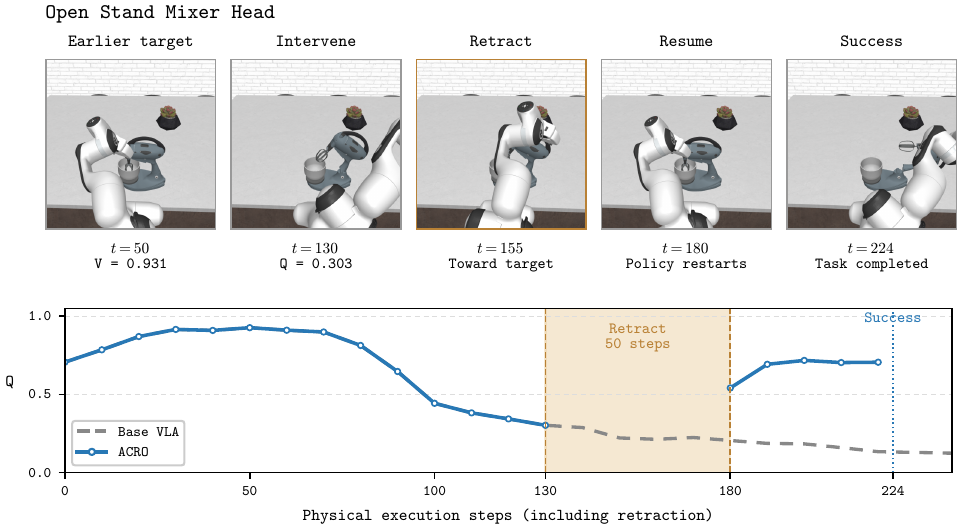}
\caption{\textbf{Retraction enables a successful re-approach.} ACRO intervenes at step 130, spends 50 steps returning toward the target at step 50, and succeeds at step 224; paired Base VLA times out at 1,000. Q curves show the first 240 steps, with no queries during retraction. Frames replay recorded actions, verified against per-step success flags.}
\label{fig:qualitative-recovery}
\end{figure}

\paragraph{Recovery through a New Approach.}
\acrorevision{Figure~\ref{fig:qualitative-recovery} shows how a return can lead to a successful continuation on Open Stand Mixer Head. ACRO intervenes at step 130 as estimated continuation success declines, then spends 50 steps returning toward an earlier target. The policy resumes and opens the mixer head after another 44 steps. Base VLA continues with low Q estimates and fails within 1,000 steps. In this example, returning to a previously visited state allows the frozen policy to make a new approach and complete the task.}

\paragraph{Execution Efficiency after Returning.}
Returning moves the robot to an earlier point along its path. If the policy then needs fewer execution steps to finish, that return has provided a more productive starting point for the next attempt. We examine the 46 trials in which both Base and ACRO succeed after ACRO's first-alarm step. From that step onward, Base uses 327.9 policy steps on average and ACRO uses 287.3, counting every retry (Table~\ref{tab:recovery-summary}). The reduction comes from open/close and pick-and-place tasks; control tasks use more policy steps under ACRO. Including return motion gives ACRO 339.9 physical steps. The lower policy-step count shows how execution after returning can reach success with fewer policy execution steps, while the physical total accounts for the movement needed to reach the re-entry state.

\Needspace{24\baselineskip}
\subsection{Test-Time Scaling and Retraction Cost}
\label{sec:data-retraction}
ACRO uses additional execution time for renewed attempts. We examine how task completion scales with this budget and how Trajectory Retraction reduces the steps spent returning between attempts.

\begin{wrapfigure}{r}{0.46\linewidth}
\vspace{-12pt}
\captionsetup{justification=raggedright,singlelinecheck=false}
\centering
\includegraphics[width=0.94\linewidth]{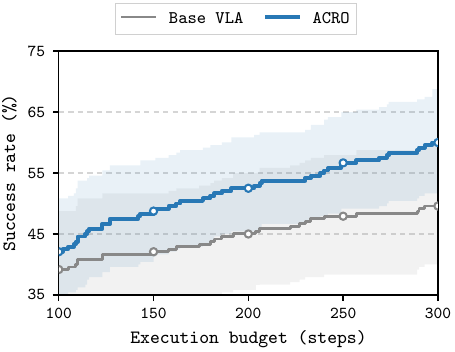}
\caption{\textbf{Success grows with budget.} Bands: 95\% bootstrap intervals.}
\label{fig:budget-curves}
\end{wrapfigure}

\paragraph{Test-Time Scaling.}
Additional execution time gives ACRO more opportunities to return and try again. We examine test-time scaling by increasing the physical execution budget and measuring how many additional tasks are completed. Figure~\ref{fig:budget-curves} scores the same 240 SimplerEnv executions at successive step limits up to 300, including all retraction steps. Within the first 100 steps, ACRO and Base VLA complete 101 and 94 trials. Over the remaining 200 steps, ACRO adds 43 completions and Base adds 25. On Eggplant in Basket, Base adds no success after step 100 while ACRO adds five; task-level curves appear in \cref{fig:budget-curves-tasks}. Thus, ACRO converts additional execution budget into further successes through renewed attempts.
\finishwrap[3]

\Needspace{14\baselineskip}
\begin{wraptable}{r}{0.49\linewidth}
\vspace{-4pt}
\captionsetup{justification=raggedright,singlelinecheck=false}
\centering
\small
\setlength{\tabcolsep}{3pt}
\caption{\textbf{Shortcuts reduce return cost.} Paths at $2H_{\mathrm{ref}}$: success (\%), mean return steps, and target reach (\%).}
\label{tab:rewind-path}
\begin{tabular}{@{}lrrr@{}}
\toprule
Path & Success & Steps & Reach \\
\midrule
Reverse replay & 70.7 & 62.0 & 86.2 \\
RDP-simplified & 70.9 & 43.4 & 80.5 \\
Straight line & 70.7 & 20.6 & 64.1 \\
\rowcolor{linkblue!12}
Ours & 71.3 & 32.4 & 85.4 \\
\bottomrule
\end{tabular}
\end{wraptable}

\paragraph{Retraction Path.}
Replaying the recorded path spends steps on detours made during the original attempt. We compare reverse replay with an RDP-simplified path, a shortcut within a tube around visited positions, and a straight line to the target. All variants use three RoboCasa tasks with 150 trials per task at $2H_{\mathrm{ref}}$ (\cref{app:protocol}). Table~\ref{tab:rewind-path} shows that the visited-tube shortcut reduces the mean return from 62.0 to 32.4 steps while retaining similar target reach (86.2\% versus 85.4\%). A straight line is shorter (20.6 steps), but reaches the target in only 64.1\% of returns. Task success stays between 70.7\% and 71.3\% across paths. The visited-tube shortcut therefore saves return steps while preserving reliable access to the target.
\finishwrap

\section{Conclusion}

We introduced ACRO, an Actor--Critic rollout orchestration layer that improves task completion by controlling when to intervene and where a fixed VLA resumes execution. The Critic guides recovery toward a promising continuation, and Trajectory Retraction makes the physical return efficient by using the visited path. Across two VLA backbones, simulation benchmarks, and real-robot manipulation, ACRO improves success under the same physical budget, including return motion. Additional execution time enables further successes through renewed attempts. These findings show that evaluating and redirecting an ongoing execution can make better use of a learned policy's capabilities, connecting execution-time orchestration to more reliable task completion.

\clearpage
\section*{AI Use Statement}

We used ChatGPT to assist with manuscript drafting and revision, discussion of mathematical formulations, literature discovery, and code for figures and tables. We did not use ChatGPT for method or experimental design, implementation or debugging of research code, or interpretation of experimental results. The authors reran the experiments to check the reported results and take responsibility for the final content of this work, including all text, claims, and artifacts produced with AI assistance.

\section*{Ethics Statement}

This study did not involve human-subject research or the collection of personal data. Real-robot experiments were conducted in a separated workspace under researcher supervision, with an emergency stop immediately accessible to the supervising researcher.

\section*{Reproducibility Statement}

Section~\ref{sec:exp_setup} describes the policies, environments, baselines, and execution budgets used in our experiments. Appendix~\ref{app:details} provides the evaluation protocol, Critic data splits and training settings, intervention rules, retraction procedure, and baseline implementations. Appendix~\ref{app:budget-curves} describes the task-level budget analysis and resampling procedure. Code is not included with this submission. Upon acceptance, we plan to release the code, trained Critic checkpoints, and experimental configurations to support reproduction of the results.

\bibliographystyle{acro_preprint}
\setlength{\bibsep}{4pt plus 1pt minus 1pt}
\bibliography{references}

\clearpage
\appendix
\section{Implementation Details}
\label{app:details}

This section describes the evaluation protocol, the components of ACRO, and our instantiation of the baselines. \Cref{tab:app-hparams} summarizes the hyperparameters.

\subsection{Evaluation Protocol}
\label{app:protocol}

\paragraph{Pairing.}
For each trial, methods use the same task, environment seed, and policy noise seed. The RoboCasa analysis retains all 600 task--seed identities; initial-observation and action-prefix checks are reported in Appendix~\ref{app:recovery-timing}. \acrorevision{Action selection can change a baseline trajectory from its first policy query; matching initial conditions does not imply identical action prefixes.} Budgets count physical steps, including the steps spent moving the robot during an intervention. Additional policy or VLM queries are not charged.

\paragraph{SimplerEnv Bridge.}
We use the official GR00T N1.7 SimplerEnv Bridge checkpoint and its four standard tasks. The initial configurations of each task, defined by object placement and orientation, are split into disjoint sets for Critic training, Critic validation, threshold calibration, and evaluation, in proportion $5{:}2{:}2{:}3$; all orientations of a placement fall in the same set. We evaluate on six held-out configurations per task, with ten trials per configuration (60 trials per task, 240 in total). For Eggplant in Basket, which has nine held-out configurations, we use the first six. The policy is queried every 4 steps and executes the first 4 steps of each 16-step action chunk. The budget is 300 steps, the limit of the official GR00T evaluation.

\paragraph{RoboCasa.}
We use the $\pi_{0.5}$ RoboCasa checkpoint released with LeRobot and 12 atomic tasks, evaluated on 50 environment seeds each (600 trials in total). Critic training uses a disjoint range of environment seeds. The evaluation budget $H$ is each task's registered episode limit, listed in \cref{tab:app-robocasa}. RoboCasa and the Bridge tasks other than Eggplant in Basket were run with a longer budget and scored at the standard limit. We retain the original fixed intervention schedules and score the saved per-step success traces up to $H$, including all executed recovery motion.

\paragraph{Analysis and ablations.}
The Critic and path ablations use three RoboCasa tasks, one per category: Turn On Electric Kettle, Open Stand Mixer Head, and Counter to Cabinet, each on 150 environment seeds disjoint from the Critic's (450 paired trials per variant). The reference horizon $H_{\mathrm{ref}}$ of each task is the 95th percentile of the lengths of successful Base VLA executions on the Critic's seeds, rounded up to a multiple of 50 steps (550, 500, and 850 steps), and every variant has a budget of $2H_{\mathrm{ref}}$ (1,100, 1,000, and 1,700 steps). Since a variant produces the same actions as Base VLA until its first intervention, we replay its trigger on the recorded Base VLA executions and simulate only the trials in which it fires; the other trials keep the Base VLA outcome.

\begin{table}[!t]
\centering
\small
\caption{RoboCasa tasks and their registered episode limits.}
\label{tab:app-robocasa}
\begin{tabular}{@{}llr@{}}
\toprule
Category & Task & Limit (steps) \\
\midrule
Open/Close/Slide & Close Fridge & 1200 \\
 & Open Cabinet & 1400 \\
 & Open Drawer & 1000 \\
 & Open Stand Mixer Head & 600 \\
 & Slide Dishwasher Rack & 600 \\
\midrule
Pick-and-Place & Counter to Cabinet & 1000 \\
 & Counter to Stove & 800 \\
 & Drawer to Counter & 1000 \\
 & Sink to Counter & 1200 \\
 & Toaster to Counter & 800 \\
\midrule
Control Operation & Turn On Electric Kettle & 600 \\
 & Turn On Sink Faucet & 800 \\
\bottomrule
\end{tabular}
\end{table}

\subsection{Critic}
\label{app:critic}

\paragraph{Data.}
The Critic is trained on uninterrupted rollouts of the base policy: 1{,}230 on SimplerEnv (780 for training, 150 for validation, 300 for calibration) and 1{,}200 on RoboCasa (720, 180, and 300). No rollout from the evaluation placements or seeds is used.

\paragraph{Architecture.}
At each query, the input is the mean-pooled output of the VLA backbone concatenated with the robot's proprioceptive state. A linear projection with layer normalization embeds each of the last eight inputs, and a single-layer GRU summarizes them. The V head predicts the value of the current state from this summary. The Q head adds a residual, computed from the summary and a projection of the proposed action chunk, to the V prediction. Both heads output hazard rates of a piecewise-exponential survival model over the next steps, with bin edges at $\{0, 10, 25, 50, 100, 200, 400, 600\}$, so that the success probability can be read at any horizon.

\paragraph{Training and selection.}
We train with AdamW (learning rate $10^{-3}$, weight decay $10^{-3}$, batch size 32, gradient clipping at 5) on the censored negative log-likelihood of both heads, for up to 120 epochs with early stopping after 12 epochs without improvement. We train eight variants (MLP or GRU summary, width 64 or 128, input noise 0 or 0.03, two seeds) and select the one with the lowest validation likelihood loss. The selected Critic is a GRU of width 64 on SimplerEnv and of width 128 on RoboCasa. On validation rollouts, the minimum Q over a rollout separates eventual successes from failures with an AUROC of 0.94 and 0.97, respectively.

\subsection{Intervention Trigger}
\label{app:trigger}

Rather than thresholding a single Q value, ACRO scores the last eight Q values with a logistic regression that predicts whether the execution will fail. The regression is fit on the validation rollouts with an L2 penalty of $10^{-3}$. Its threshold $\tau$ is set on the calibration rollouts as the $(1-\alpha)$ quantile of the maximum score reached by each successful rollout, with $\alpha = 0.2$, so that about one in five executions that would succeed raises an alarm. On the calibration rollouts, the trigger detects 76\% of failures on SimplerEnv and 85\% on RoboCasa. After an intervention, the trigger re-arms once two consecutive queries rise at least 0.02 above the threshold, or after 16 consecutive queries below it. The Critic is queried at every policy query on SimplerEnv and every 10 steps on RoboCasa.

\subsection{Retraction and Re-entry}
\label{app:retraction}

\paragraph{Target selection.}
The $k$-th intervention in an execution searches a look-back window of the executed path and moves the robot to the state with the highest predicted value in that window. The windows grow with $k$: 24 steps, 60 steps, and then the start of the execution on SimplerEnv, and 120 steps, 300 steps, and then the start on RoboCasa.

\paragraph{Path.}
Trajectory Retraction replaces the recorded end-effector path with straight segments that stay within 2\,cm of positions the robot has already visited, which removes loops and hesitations while staying in explored space. The robot follows this path with end-effector pose commands, clipped per step to 3\,cm and 0.3\,rad on SimplerEnv and to 5\,cm and 0.5\,rad on RoboCasa. The gripper keeps its last commanded state, the mobile base and torso stay fixed, and objects are not reset.

\paragraph{Limits.}
On RoboCasa, ACRO intervenes at most three times per execution, and a single retraction is capped at 200 steps. On SimplerEnv, the number of interventions is bounded only by the budget. \acrorevision{The action queue is cleared after retraction so that the next proposal uses the reached observation.}

\subsection{Baselines}
\label{app:baselines}

\paragraph{Periodic rewind.}
This baseline uses a reference horizon $H_{\mathrm{ref}}$ fixed from development rollouts. At each 10-step query, it triggers once $\lfloor H_{\mathrm{ref}}/2\rfloor$ physical steps have elapsed since the preceding return ended, or since episode start for the first return. It selects the recorded position $\lfloor H_{\mathrm{ref}}/4\rfloor$ logical forward steps back, bounded by episode start, and follows the recorded path in reverse. The original intervals and offsets are held fixed when scoring at the standard evaluation budget $H$. In the task order of \cref{tab:app-robocasa}, the intervals are 500, 500, 250, 250, 150, 425, 325, 475, 375, 325, 275, and 200 steps. Its trigger, target, and path are fixed rules, so the comparison measures the full recovery strategy. The policy weights and physical-step accounting are shared with ACRO.

\paragraph{RoboMonkey.}
At each policy query, we draw eight action chunks: the chunk that Base VLA would execute and seven more sampled with different noise seeds. The ACRO Critic scores each chunk as the Q value of executing it from the current state history, and the robot executes the chunk with the highest score. On SimplerEnv, a sampled chunk replaces the default only if its score is strictly higher. The original RoboMonkey trains a VLM-based verifier on preference data. Using our Critic instead lets RoboMonkey and ACRO rely on the same value estimates. RoboMonkey queries the policy eight times as often as Base VLA.

\paragraph{CycleVLA.}
A VLM (Qwen3.8-Flash) decomposes each instruction once into a short sequence of subtasks, using a fixed vocabulary of verbs (move, rotate, open, close, grasp, lift, place, release). We freeze these decompositions before evaluation, so every trial of a task uses the same subtasks. A change in the sign of the gripper command that persists for two steps marks the end of a subtask. At the next policy query, the VLM receives the instruction, the subtask list, and the current camera image, and returns whether the subtask succeeded and, if not, which subtask to resume from. On a failure, the robot moves back along its recorded path to the start of that subtask and the policy continues from there. Each subtask is retried at most twice and is then treated as complete. If the VLM does not return a valid answer, execution continues without intervention. Unlike the original work, we do not fine-tune the policy with a progress head, since every method keeps the base policy fixed. The SimplerEnv decompositions are: Carrot on Plate, (grasp carrot, move carrot to plate, place carrot on plate); Spoon on Towel, (grasp the spoon, place the spoon on the towel, release the spoon); and four subtasks each for Eggplant in Basket and Stack Cube.

\begin{table}[!t]
\centering
\small
\caption{Hyperparameters of ACRO and the baselines.}
\label{tab:app-hparams}
\begin{tabular}{@{}lcc@{}}
\toprule
 & SimplerEnv & RoboCasa \\
\midrule
Base policy & GR00T N1.7 & $\pi_{0.5}$ \\
Step budget & 300 & registered limit (\cref{tab:app-robocasa}) \\
\midrule
\multicolumn{3}{@{}l}{\emph{Critic}} \\
Rollouts (train / validation / calibration) & 780 / 150 / 300 & 720 / 180 / 300 \\
Summary network & GRU, width 64 & GRU, width 128 \\
History length & 8 queries & 8 queries \\
Query interval & 4 steps & 10 steps \\
Optimizer & \multicolumn{2}{c}{AdamW, lr $10^{-3}$, weight decay $10^{-3}$, batch 32} \\
\midrule
\multicolumn{3}{@{}l}{\emph{Trigger}} \\
History length & 8 Q values & 8 Q values \\
$\alpha$ & 0.2 & 0.2 \\
$\tau$ & 2.11 & 1.48 \\
\midrule
\multicolumn{3}{@{}l}{\emph{Retraction}} \\
Look-back windows & 24, 60, start & 120, 300, start \\
Path tube radius & 2\,cm & 2\,cm \\
Max.\ interventions & -- & 3 \\
\midrule
\multicolumn{3}{@{}l}{\emph{Baselines}} \\
RoboMonkey candidates & 8 & 8 \\
CycleVLA VLM / retries & Qwen3.8-Flash / 2 & Qwen3.8-Flash / 2 \\
\bottomrule
\end{tabular}
\end{table}

\FloatBarrier
\section{Ablation Studies}
\label{sec:ablations}
\paragraph{Overview.}
We vary the Critic architecture and intervention trigger, changing one component at a time while keeping the rest at the default configuration. All variants use the same three RoboCasa analysis tasks and $2H_{\mathrm{ref}}$ budgets as the Critic training-data experiment in Section~\ref{sec:main_results} and the retraction experiment in \cref{sec:data-retraction}.

\begin{table}[!t]
\centering
\small
\caption{Critic architecture. The MLP and KAN read only the latest query; the GRU and Transformer read the last eight. Success rate (\%), averaged over three tasks.}
\label{tab:critic-arch}
\begin{tabular}{@{}lccc@{}}
\toprule
Critic & Params & Q AUROC & Success \\
\midrule
\rowcolor{linkblue!12}
GRU (ours)  & 0.44M & 0.897 & 71.3 \\
MLP         & 0.35M & 0.886 & 71.1 \\
KAN         & 1.17M & 0.939 & 77.6 \\
Transformer & 0.62M & 0.915 & 73.6 \\
\bottomrule
\end{tabular}
\end{table}

\noindent\textbf{Critic Architecture.}
We also test whether ACRO depends on a particular Critic design. We replace the GRU with an MLP and a KAN that read only the latest query, and with a Transformer over the same eight-query window, all reading the same policy features. \Cref{tab:critic-arch} shows that every Critic improves over Base VLA and that success follows the Critic's AUROC. The MLP matches the GRU (71.1\% versus 71.3\%), the Transformer reaches 73.6\%, and the KAN attains both the highest AUROC (0.939) and the highest success (77.6\%), gaining 63 trials and losing 35 relative to the GRU. \acrorevision{ACRO improves success with every tested architecture; across these fits, higher Critic AUROC is associated with higher task success.}

\begin{figure}[!t]
\centering
\includegraphics[width=\linewidth]{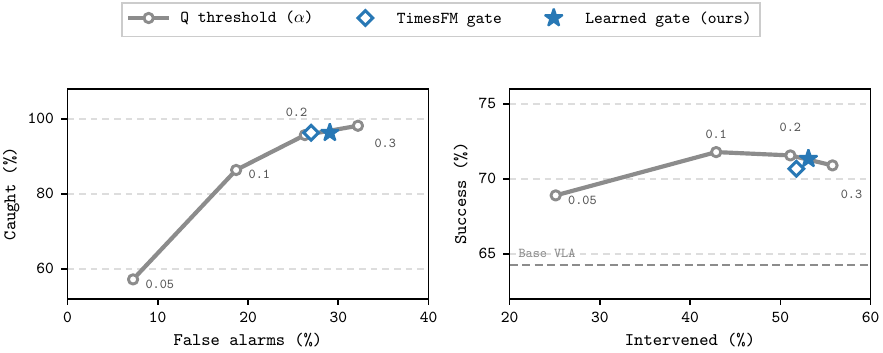}
\caption{\textbf{Intervention trigger.} Gray points threshold Q at conformal level $\alpha$; both gates are calibrated at $\alpha=0.2$. Failures caught and false alarms are the fractions of executions that fail or succeed under Base VLA in which the trigger intervenes. Averaged over three tasks (450 paired trials).}
\label{fig:trigger}
\end{figure}

\noindent\textbf{Intervention Trigger.}
Finally, the trigger trades missed failures against unnecessary interruptions of executions that would have succeeded. We vary the conformal level $\alpha$ of a Q threshold and compare it with our learned gate on the recent Q history and with a gate that forecasts Q using TimesFM. \Cref{fig:trigger} plots the fraction of failures each trigger catches against its false-alarm rate, and success against how often it intervenes. The learned gate (96.3\% with 29.1\%) and the TimesFM gate (96.3\% with 27.0\%) fall on the same trade-off as the threshold at $\alpha=0.2$ (95.7\% with 26.3\%). Success saturates once most failures are caught: every trigger with $\alpha\ge0.1$ reaches 70.7\% to 71.8\%, whereas $\alpha=0.05$ reaches 68.9\%. Thus, on these tasks, how many failures the trigger catches matters more than whether it reads the level or the trend of Q.

\FloatBarrier

\section{Success over the Execution Budget}
\label{app:budget-curves}
\begin{figure}[!t]
\centering
\includegraphics[width=\linewidth]{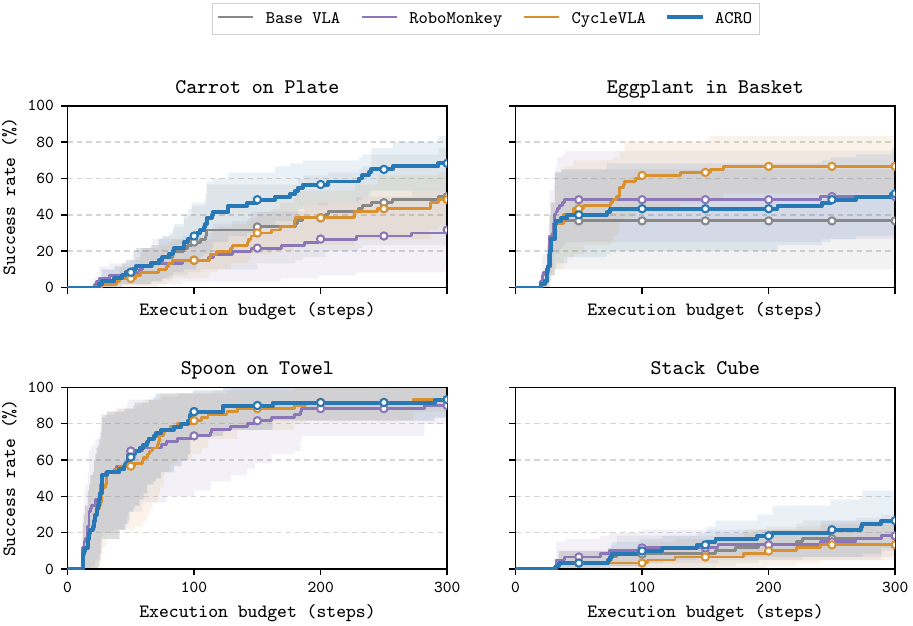}
\caption{\textbf{Success over the execution budget by task.} Sixty paired trials per task on SimplerEnv Bridge, comparing Base VLA, RoboMonkey, CycleVLA, and ACRO. All panels use the same success-rate scale. Bands show 95\% bootstrap intervals over object placements.}
\label{fig:budget-curves-tasks}
\end{figure}

The curves in \cref{fig:budget-curves-tasks} disaggregate \cref{fig:budget-curves} by task. Every method is evaluated on the same 60 trials per task and all physical movement counts toward the 300-step budget. The shaded intervals resample object placements, retaining all ten trials of each sampled placement.

\FloatBarrier
\section{Recovery Outcomes and Execution Cost at the Standard Budget}
\label{app:recovery-timing}

\paragraph{Cohort and budget.}
This analysis uses all 600 RoboCasa task--seed identities from the main evaluation: 12 tasks and 50 environment seeds per task. Each execution is scored at the task's registered physical-step limit $H$ in \cref{tab:app-robocasa}. Success is read from the saved per-step environment success array. All 3,000 records across the five methods cover their scoring horizon or succeed before it, and their saved terminal success flags agree with the full success arrays. The source manifest records commit \texttt{f2493a4} and the frozen \texttt{lerobot/pi05\_robocasa} policy. The original fixed schedules are preserved during prefix scoring. The separate three-task ablations use their original longer budgets.

\paragraph{Outcome accounting.}
ACRO completes 408 trials, Base 364, and Periodic rewind 357. Relative to Base, ACRO gains 70 successes and loses 26; Periodic gains 50 and loses 57. Thus, the net changes are $+44$ and $-7$. The ACRO comparison contains 338 shared successes and 166 shared failures. These four outcome groups sum to 600 without excluding any identity.

\paragraph{Pairing checks.}
Task and policy seeds are shared, but exact observations and action prefixes are not identical for every execution. Compared with Base, 45 ACRO records and 60 Periodic records have different initial-observation hashes. Action arrays before the first changed action differ for 57 ACRO and 75 Periodic records. The full 600-identity outcome comparison retains these records and describes observed same-seed differences.

\paragraph{Timing after the first alarm.}
ACRO initiates a return before success or cutoff on 284 of the 600 identities. Of the 338 shared successes, 46 succeed in both executions after the step index of ACRO's first alarm. We use that index as the time origin and subtract all ACRO return actions before success to obtain policy execution steps. Every retry remains in the count. This gives 327.9 Base policy steps and 287.3 ACRO policy steps on average; including return motion gives 339.9 ACRO physical steps. The median paired policy-step difference is $-5$ steps. Four of these 46 records have different initial-observation hashes and five have different action prefixes, so the timing comparison is descriptive. It conditions on shared success and uses no exclusion based on the pairing checks.

\begin{table}[!t]
\centering
\small
\caption{Shared successes after ACRO's first alarm at $H$. Means count every policy attempt; ACRO physical steps also include return motion.}
\label{tab:paired-recovery}
\begin{tabular}{@{}lrrrr@{}}
\toprule
 & & \multicolumn{2}{c}{Policy steps} & Physical steps \\
\cmidrule(lr){3-4}\cmidrule(l){5-5}
Category & $N$ & Base & ACRO & ACRO \\
\midrule
Open/Close/Slide & 20 & 362.6 & 296.1 & 365.2 \\
Pick-and-Place & 17 & 387.0 & 345.5 & 391.2 \\
Control & 9 & 139.3 & 157.7 & 187.0 \\
\midrule
Overall & 46 & 327.9 & 287.3 & 339.9 \\
\bottomrule
\end{tabular}
\end{table}

\paragraph{Physical execution cost.}
Return actions count against $H$. For each return event, we intersect the interval from its first action through its last executed action with the interval before the first success or $H$. This also includes unfinished returns at the cutoff. Averaged over all 600 trials, Periodic initiates 1.04 returns and spends 92.7 steps returning; ACRO initiates 1.06 and spends 46.3 steps returning. Total executed physical steps average 553.6 for Base, 595.9 for Periodic, and 528.4 for ACRO. These totals include successes and failures; they describe physical execution, without measuring inference latency or wall-clock duration.

\FloatBarrier

\end{document}